# A Recursive Robust Filtering Approach for 3D Registration

Abdenour Amamra, Nabil Aouf, Dowling Stuart and Mark Richardson

*Centre for Electronic Warfare, Cranfield University, United Kingdom*

{a.amamra, n.aouf, s.dowling, m.a.richardson}@cranfield.ac.uk

## Abstract

This work presents a new recursive robust filtering approach for feature-based 3D registration. Unlike the common state-of-the-art alignment algorithms, the proposed method has four advantages that have not yet occurred altogether in any previous solution. For instance, it is able to deal with inherent noise contaminating sensory data; it is robust to uncertainties caused by noisy feature localisation; it also combines the advantages of both $L_\infty$ and $L_2$ norms for a higher performance and a more prospective prevention of local minima. The result is an accurate and stable rigid body transformation. The latter enables a thorough control over the convergence regarding the alignment as well as a correct assessment of the quality of registration. The mathematical rationale behind the proposed approach is explained and the results are validated on physical and synthetic data.

*Keywords: time-varying 3D Registration; Recursive Least Squares; Kalman Filter; Robust $H_\infty$ Filter.*

## Introduction

The widespread abundance of affordable 3D sensing devices has encouraged many enthusiasts to contribute new solutions for 3D reconstruction [1]. The latter require data alignment tools that enable the recovery of the 6DOF regarding viewpoints where the different scans had been captured. Theoretically, each viewpoint has a different coordinate system. Knowledge of the transformation that maps a given 3D point from one frame to another, therefore, becomes necessary.

In practice, the alignment requires some keypoints from a *Source* and a *Target* point cloud. Hence, alignment problem amounts to the determination of the mapping between the source and the target frames. To this end, we assume the keypoints being available and we focus on 3D registration. Generally, the determination of the best transformation is based on $L_2$ norm minimisation. However, $L_2$ optimisers assume a prior availability of the entire datasets before processing takes place. From a practical point of view, such an assumption is too optimistic due to sizeable noisy data streamed at relatively high frame rates that one encounters in practice. For this reason, our novel 3D registration solution delivers the 6DOF pose between viewpoints recursively and is capable of handling 3D points' noise and uncertainty for a more efficient estimation.

The remainder of this paper is organised as follows: In the first section, the related works about 3D registration are discussed and different alignment solutions that had been proposed so far are analysed. In the following section, 3D registration problem is formulated in a Least Squares (LS) form. In the next section the link between 3D registration and RLS is settled and fitted into Kalman filter's (KF) equations [2]. The parametric uncertainty of the 3D feature points is afterwards determined to be later used in the Robust $H_\infty$ (RF) modelling for sparse alignment. Our contribution is validated on both synthetic and real datasets. Lastly, the paper is concluded and potential future works are recommended.

## Related works

Since its invention by Besl et al. [3], the Iterative Closest Point algorithm (ICP) has been considered as a reference in point cloud alignment literature. However, a good initial guess or some feature correspondences are necessary to avoid local minima. Newer variants of that algorithm have been proposed to deal with its limitations such as EM-ICP [4] and Softassign [5]. Unlike the original implementation that assigns to every point in the source its closest correspondent in the target, subsequent variants allow each point to be checked against the entire target dataset. To this end, weighting coefficients are associated with the elements to discard the describe their quality [5]. Other variants inspired by the original algorithm (ICP) were further proposed such as non-linear ICP [6], generalised ICP [7] as well as non-rigid ICP [8]. Larusso et al. [9] showed that all closed-form solutions are computationally similar. However, performance can significantly differ from one solution to another. Thus, no single algorithm is exclusively optimal for all scenarios. Umeyama [10] states in his work that Horn and Arun's algorithms fail when the datasets become highly corrupted with noise. He further proposed an alternative solution that utilises Lagrange Multipliers [11].

A solution for the recursive estimation of rigid body transformations with the Extended Kalman Filter (EKF) was first proposed by Pennec et al. [12]. Ma et al. [13] followed the same strategy in order to align datasets contaminated with isotropic Gaussian noise using the Unscented Particle Filter (UPF) [14]. This algorithm can accurately estimate the parameters for very small datasets (less than one hundred elements). An Unscented Kalman Filter (UKF) algorithm was also adapted by Julier et al. [15] to align two datasets following a sequential estimation. All these recursive algorithms minimise $L_2$ norm but consider the parameters being accurately determined beforehand. Nevertheless, it is impossible to assert the certainty of parameters in real scenarios. In our solution, however, we consider them (parameters) being uncertain, and we confine estimation error to a small range by optimising $L_\infty$ norm instead of $L_2$.

Micusik et al. [16] used a Second Order Cone Programming (SOCP) to minimise the $L_\infty$ norm for non-overlapping cameras. They have shown a good performance with a fairly small error magnitude. Lee et al. [17] further claimed that by using $L_\infty$ a number of computer vision problems such as homography estimation can be formulated and solved using Bisection method.

In the light of this background, our work takes advantage of the mature recursive estimation framework in order to compute a robust and optimal solution for 3D registration problem by means of $L_\infty$ norm minimisation.

## Problem statement

Given two sets of source and target 3D point clouds $Q = \{q_1, \ldots, q_n\}$, $P = \{p_1, \ldots, p_n\}$ respectively. Each of the elements $p_i, q_i$ within the sets of points has three components $p_i = (x_p, y_p, z_p)_i$ and $q_i = (x_q, y_q, z_q)_i$. The $k$-th point $q_k$ in the source point cloud has been matched a priori with the $k$-th point in the target point cloud $p_k$. The purpose of 3D registration is to find a rigid body transformation ($R$: rotation, $t$: translation) that maps the source $Q$ onto the target $P$. The determination of such a mapping can be modelled as an optimisation problem [18]. Nevertheless, due to noisy outputs streamed by the sensor, an exact solution is very unlikely to determine. Thus, a realistic model must take into account alignment error $e_i$ as follows:

$$p_i = Rq_i + t + e_i \tag{1}$$

The rigid body transformation $[R, t]$ is optimal when the sum of the squares of errors ($e_i$) becomes minimal:

$$e^2 = \underset{R,t}{argmin} \sum_{i=1}^{n} \|p_i - (Rq_i + t)\|^2 \tag{2}$$

Where:

$$R = \begin{bmatrix} r_{11} & r_{12} & r_{13} \\ r_{21} & r_{22} & r_{23} \\ r_{31} & r_{32} & r_{33} \end{bmatrix} ; \; t = \begin{bmatrix} t_x \\ t_y \\ t_z \end{bmatrix} \tag{3}$$

It is possible to simplify the problem of equation (2) by decoupling the translation vector $t$ and eliminating scale difference as follows:

$$\hat{t} = \bar{p} - \hat{R}\bar{q} \; ; \quad \hat{s} = \sum_{i=1}^{n} \left( \frac{\bar{p}_i \hat{R} \bar{q}_i}{\|\bar{q}_i\|^2} \right) \tag{4}$$

$$\tilde{t} = \bar{p} - \bar{q} \; ; \quad \tilde{s} = \sum_{i=1}^{n} \left( \frac{\bar{p}_i \bar{q}_i}{\|\bar{q}_i\|^2} \right) \tag{5}$$

As claimed by Horn [19], $\bar{q}$ and $\bar{p}$ are the centroids respective to the source and the target point clouds; $\hat{t}, \hat{s}$ are the optimal translation and scale between the two dataset. Whereas, $\tilde{t}, \tilde{s}$ are their respective initial guesses when the initial rotation is assumed to be $R_0 = I_3$. As a result of this simplification, the problem of pose estimation in equation (2) is now reduced to:

$$e^2 = \underset{R}{argmin} \sum_{i=1}^{n} \|\bar{p}_i - R\bar{q}_i\|^2 \tag{6}$$

Once the optimal rotation $\hat{R}$ computed, $\hat{t}$ and $\hat{s}$ can be deduced using equation (4). On the other hand, the optimal rotation $\hat{R}$ can be obtained by minimising $\sum_{i=1}^{n} \|\bar{p}_i - R\bar{q}_i\|^2$ using a LS optimiser. The resulting estimation is sufficient for most applications as long as robustness is not a determining factor. However, if the inputs become significantly contaminated with noise, the result becomes unstable (i.e. very sensitive to perturbations in the data) and more likely to drift away from the optimal solution.

## 3D Registration with RLS

Despite the performance of time-varying filters, 3D registration has profited very poorly from their assets even after closed-form methods were proven weak in various practical situations. Moreover, the authors of a number of recent image registration surveys did not even allude to the possibility of solving 3D alignment with recursive filtering tools [20]. The power of the recursive solutions can be appreciated due to what has been claimed earlier and to the possibility of cooperation between different registration instances working together. The latter can share their most updated estimates instantaneously. As a result, they can benefit from each other's contributions, which in turn reduces the probability of falling into a local minimum.

### Recursive Modelling of 3D Registration

In order to express the cost function of equation (6) in a recursive fashion, the original problem should be rewritten as shown in equations (7) to (10). Such a transformation allows us to fit 3D registration problem in a recursive least squares framework.

$$p = Rq + e \tag{7}$$

$$\begin{cases} p_x = r_{11}q_x + r_{12}q_y + r_{13}q_z + e_x \\ p_y = r_{21}q_x + r_{22}q_y + r_{23}q_z + e_y \\ p_z = r_{31}q_x + r_{32}q_y + r_{33}q_z + e_z \end{cases} \tag{8}$$

By analogy, the state variable $x_k$ now represents the rotation matrix $R$ of equation (7). The optimiser uses pairs of corresponding points in order to refine the entries of the state vector now containing the entries of rotation matrix $\mathcal{R}_9$. For instance, at every time-step $k$ we have:

$$v = [q_x \quad q_y \quad q_z] \tag{9}$$

$$\begin{bmatrix} p_x \\ p_y \\ p_z \end{bmatrix} = \begin{bmatrix} v & & \\ & v & \\ & & v \end{bmatrix} \begin{bmatrix} r_{11} \\ r_{12} \\ r_{13} \\ r_{21} \\ r_{22} \\ r_{23} \\ r_{31} \\ r_{32} \\ r_{33} \end{bmatrix} + \begin{bmatrix} e_x \\ e_y \\ e_z \end{bmatrix}$$

$$p = H_R \mathcal{R}_9 + e \quad (10)$$

$x_k = [r_{11}\ r_{12}\ r_{13}\ r_{21}\ r_{22}\ r_{23}\ r_{31}\ r_{32}\ r_{33}]^T \in \Re^9$; $A_k = I_9$; $B_k = 0_9$ as no control variable is required. $w_k \sim \mathcal{N}(0, Q_k)$ is a random variable representing process noise for which $Q_k = \sigma_k^2 I_9$; $\sigma_k > 0$ should be small because the process is accurately determined. $z_k \in \Re^3$ is the actual noisy measurement vector whose elements are the coordinates of the target feature point. $y_k \in \Re^3$ is the predicted observation vector that contains the 3D position of the target feature point. $v_k$ is a random variable for which $R_k = [\sigma_x\ \sigma_y\ \sigma_z]I_3$, it represents noise process contaminating target feature point localisation. The complete scheme of KF-based registration is explained in Algorithm 1. The latter works as follows: 1) Initialise the state vector (rotation matrix) with the entries of $I_9$. If available, an initial guess would be preferable. 2) Iterate over feature points; acquire a new target feature $z_k$ and build $H_k$. 3) KF prediction. 4) KF correction where the estimate $x_k$ and the covariance of error in estimation $P_k$ are corrected with $K_k$.

Algorithm 1 KF-based registration

*Source and target point clouds*
**P, Q**: *3D feature points;*
$[p_i, q_i] = \emptyset$: *Correspondences list;*
$[p_i, q_i] = FindCorrespondences(P,Q);$
$\hat{x}_0 = [1.0\ 0.0\ 0.0\ 0.0\ 1.0\ 0.0\ 0.0\ 0.0\ 1.0]^T$
$Q = \sigma I_9$
$\hat{P}_0 = Q$
$R_0 = [\sigma_x\ \sigma_y\ \sigma_z]I_3$
*for each pair of correspondences* $(k = 1, n)$
  $z_k = [\ p(k).x\ \ p(k).y\ \ p(k).z]$
  $H_k = H_R(p(k))$
  *Prediction*
    $x_k = \hat{x}_{k-1}$ (11)
    $y_k = H_k x_k$ (12)
    $P_k = \hat{P}_{k-1} + Q$ (13)
  *correction*
    $K_k = P_k H_k^T (H_k P_k H_k^T + R_k)^{-1}$ (14)
    $\hat{x}_k = x_k + K_k(z_k - y_k)$ (15)
    $\hat{P}_k = (I - K_k H_k) P_k$ (16)
*end*

The computational complexity of KF registration is proportional to $O(n \times 9^3)$ in the worst case, where $n$ is the number of keypoints used to compute the optimal registration and **9** is the size of the state vector. On the other hand, the best complexity regarding alternative registration algorithms such as ICP, EMICP and WICP is proportional to $O(n^2 \times 3^{2.37})$. KF 3D registration can be easily expanded to include the three components of translation vector in $H_k$.

# Robust $H_\infty$ Registration

### 3D points uncertainty

In order to handle instability in parameters estimation, the uncertainties should be confined into a small range. To this end, the behaviour of the noisy inputs must be thoroughly studied. Uncertainties are modelled empirically by looking at how 3D points are distributed, and how do 3D sensors sense the real world.

### z-Resolution of RGBD Cameras

The authors have already shown in a previous research [21] that the points within a 3D image lie on parallel clusters that were named **"Z-Levels"**. Such a structure allows us to quantify correctly the amount of uncertainty in every feature point.

### Depth Noise Statistics

RGBD sensors' measurement-noise has a Gaussian distribution with varying standard deviations. These standard deviations rely on the range between the sensor and the scene. The standard deviation $\sigma_{z_k}$ of a given Z-level $z_k$ is defined by the length of the interval where $z_k$ is expected to vary as shown below:

$$\sigma_{z_k} = (Z_{k+i} - Z_{k-i})/2 \quad (17)$$

Here, $\sigma_{z_k}$ represents the average distance separating the two Z-levels $Z_{k+i}$ and $Z_{k-i}$ and the central one $z_k$. Empirically, the best estimation of the standard deviation regarding noise affecting the 3D points lying on $z_k$ is obtained when $i = 3$. That is, the true depth $\hat{z}_k$ taken by a given Z-level is expected to be equal to $z_k \pm ((Z_{k+3} - Z_{k-3})/2)$. The standard deviations concerning the remaining two coordinates $(x_k, y_k)$ are deduced from the intrinsic parameters of the camera $(f_x, f_y, c_x, c_y)$ and $\sigma_{z_k}$ as follows:

$$\begin{cases} u_i = (f_x/z_i)x_i + c_x \\ v_i = (f_y/z_i)y_i + c_y \end{cases} \quad (18)$$

$$\begin{cases} x_i = (z_i/f_x)(u_i - c_x) \\ y_i = (z_i/f_y)(v_i - c_y) \end{cases} \quad (19)$$

$$\begin{cases} \sigma_{z_k} = 0.5\,(Z_{k+i} - Z_{k-i}) \\ \sigma_{x_k} = (\sigma_{z_k}/f_x)(u_k - c_x) \\ \sigma_{y_k} = (\sigma_{z_k}/f_y)(v_k - c_y) \end{cases} \quad (20)$$

Every point is, therefore, affected by certain amount of noise characterised by the standard deviations $\sigma_{x_k}, \sigma_{y_k}, \sigma_{z_k}$ towards the directions of the axes $x, y$ and $z$, respectively. Hence, the covariance matrix attributed to each point $p(x, y, z)$ is described as:

$$C(x, y, z) = \begin{bmatrix} \sigma_x^2 & \sigma_x\sigma_y & \sigma_x\sigma_z \\ \sigma_y\sigma_x & \sigma_y^2 & \sigma_y\sigma_z \\ \sigma_z\sigma_x & \sigma_z\sigma_y & \sigma_z^2 \end{bmatrix} \quad (21)$$

$C$ represents the spread of uncertainty around the point $p(x, y, z)$. As can be seen in Figure 1 (a), the projection of covariance ellipsoids of a given 3D point on the planes $zx, zy, yx$ yields three ellipses. The more accurately a feature point is captured, the smaller the norm of its covariance matrix (blue point in Figure 1 (a)). On the other hand, the less accurate the capture of a given feature is, the larger the norm of its covariance matrix (red point in Figure 1 (a)).

Kanazawa et al. [22] claimed that the incorporation of feature uncertainty does not contribute any further improvements to the estimation. On the other hand, Brooks et al. [23] as well as us in a previous work [24], both noticed a reduced error in estimation after considering uncertainty. Based on the conducted experiments with registration algorithms and the fact that Weighted-ICP (WICP takes into account data uncertainty) outperforms ICP, as will be shown in the results, it is obvious that the incorporation of feature-location uncertainty improves pose estimation remarquably.

## Robust $H_\infty$ (RF) Filter for 3D Registration

In this section, we propose a time-varying registration algorithm that incorporates modelling and measurement uncertainties as follows:

$$x_k = (A_k + \Delta A_k)\hat{x}_{k-1} + B_k u_k + w_k \quad (22)$$

$$y_k = (H_k + \Delta H_k)x_k + v_k \quad (23)$$

$\Delta H_k$ represents the uncertainty in observation model, whereas $\Delta A_k$ is the uncertainty in process model. In our case, the two matrices take the values:

$$\Delta A_k = \sigma_A I_9$$
$$\sigma_A = [\sigma_{r_{11}} \sigma_{r_{12}} \sigma_{r_{13}} \sigma_{r_{21}} \sigma_{r_{22}} \sigma_{r_{23}} \sigma_{r_{31}} \sigma_{r_{32}} \sigma_{r_{33}}] \quad (24)$$
$$V_k = [q_x + \sigma_x \quad q_y + \sigma_y \quad q_z + \sigma_z]$$

$$\Delta H_k = \begin{bmatrix} V_k & & \\ & V_k & \\ & & V_k \end{bmatrix} \quad (25)$$

If these matrices cannot be determined, RF would still be able to control the instability disturbing its parameters [25] by assuming it being of the form:

$$\begin{bmatrix} \Delta A_k \\ \Delta H_k \end{bmatrix} = \begin{bmatrix} M_{1k} \\ M_{2k} \end{bmatrix} \Gamma_k N_k \quad (26)$$

$M_{1k}, M_{2k}$ and $N_k$ are known matrices, $\Gamma_k$ is unknown but it should satisfy the bound:

$$\Gamma_k^T \Gamma_k \leq I \quad (27)$$

Our purpose is to design a state estimator of the form:

$$x_{k+1} = \tilde{A}_k x_k + \widetilde{K}_k t_k \quad (28)$$

The latter should be stable (the eigenvalues of $\tilde{A}_k$ must be less than one in magnitude). The determination of the parameters of the filter can be done through the procedure described in our previous work [24].

The adaptation of RF is proven to be flexible and capable of delivering accurate state estimations, however uncertain system's parameters are. Estimation error compared to the ground truth measurements will show the effectiveness of RF 3D registration against alternative non-robust methods such as KF and the more established algorithms available in the literature. In real scenarios, the exact model is very unlikely to determine [26]. Yet the non-robust tools do not consider uncertainties in their parameters. Hence, if by chance the parameters are accurate, these tools perform as well as RF. On the other hand, when the system is not precisely characterised, they become significantly unstable. For instance, RF registration combines the robustness of $H_\infty$ (it is less affected by the accuracy of system's parameters) and the optimality of KF on linear systems to produce an accurate and stable estimate. Such a quality guaranties a high precision of estimation and more stability towards inputs' perturbations.

## Results & Discussions

In this section, the results regarding KF and RF registration are validated with tests on real and synthetic 3D data. Our test benchmark includes: WICP [27]; Expectation Maximisation ICP algorithm (EMICP)[28] and Horn's closed form solution based on quaternions (HORN)[29].

Here, accuracy is measured by the distance separating the target and the source point clouds after the registration. In order to fairly assess every algorithm, processing time elapsed to find the best pose is also recorded. Throughout experiments, it is noticeable that the plotted metrics (processing time and RMSE= $\sqrt{\frac{1}{n}\sum_{i=1}^{n}\|\hat{R}Q_i + \hat{t} - P_i\|^2}$ ) are not homogeneous. For this reason, a logarithmic scale was used to cope with the difference of scale within the same plot.

The number of keypoints extracted from every point cloud is about 400 points. In practice, an average-sized point cloud in a single frame contains up to 400 useful key points. Computation time has been calculated for the five algorithms running on an **i7-2670QM** working at **2.2GHz**, with **12.0GB** of memory. A sample is a set of 400 pairs of corresponding <*source, target*> keypoints. 30 samples were tested in each of the following five scenario (two with real and three with synthetic data).

4**Real data**

In this experiment, image data is delivered by two versions of Kinect[1] sensor (Kinect 1 is based on structured light principle; whereas, Kinect 2 is a time-of-flight camera). In addition, SIFT3D extractor and CSHOT [21] descriptor were used to obtain feature points from the real data.

In order to collect real 3D point clouds, the camera was carried and moved around in an infinity-shaped (∞) trajectory within the arena of our autonomous navigation lab. Simultaneously, a high-quality tracking system (OptiTrack[2]) was used as a ground truth reference, Figure 1 (**b**). 120 different pairs of overlapping point clouds were captured by each of the two Kinects. RGBD image data acquisition runs simultaneously as the robot moves around. At each time-step, we acquire a single pair of colour and depth images (both constitute a single point cloud) for the indoor scene. Hence, a total of 120 pairs of point clouds are aligned in a pairwise manner between ($C_i$, $C_{i+1}$). The last sample $C_{120}$ is registered against both $C_{119}$ and $C_1$ to test the loop closure.

**Scenario 1: New Kinect**
**RMSE:** The average RMSE for the five algorithms (see Figure 1 (c)) was as follows: 0.27m for EMICP (pink), 0.13m for WICP (green), 0.28m for Horn (red), 0.15mm for RF (black) and 0.7mm for KF (blue).

**Scenario 2: Old Kinect**
**RMSE** was 0.28m for EMICP, 0.22m for WICP, 0.3m for Horn, 0.95mm for RF and 1.13mm for KF (see Figure 1 (d)).
**Average processing time for both scenarios** was 114.3ms for EMICP, 26.7ms for WICP, 1.05ms for Horn, 23.1ms for RF and 11.64ms for KF (see Figure 1 (e), (f)).

**Synthetic Data**

In this experiment, we consider only artificial 3D keypoints, where, $Q_i$ (source keypoints) as well as a random 3D transformation $[R_i, t_i]$ had been generated randomly. The target 3D keypoints are built using the equation, $P_i = R_i Q_i + t_i$. To realistically simulate physical data, a normally distributed anisotropic white noise was added to the clean datasets. The latter had different magnitudes $\sigma_i$: *large* ( $20\ mm \leq \sigma_i \leq 80\ mm$), *average* ( $10\ mm \leq \sigma_i \leq 20\ mm$) and *small* ( $0.1\ mm \leq \sigma_i \leq 10\ mm$). For each, is generated 1000 point clouds, results were as follows:

**Scenario 1: Small Noise Magnitude**
**RMSE** was 0.42m for EMICP, 0.18m for WICP, 0.46m for Horn, 0.18mm for RF and 0.44mm for KF (see Figure 2 (a)).

**Scenario 2: Average Noise Magnitude**
**RMSE** was 0.54m for EMICP, 0.48m for WICP, 0.56m for Horn, 0.22mm for RF and finally, 0.43mm for KF (see Figure 2 (b)).

**Scenario 3: Large Noise Magnitude**
**RMSE** was 0.49m for EMICP, 0.35m for WICP, 0.51m for Horn, 0.63mm for RF and 0.89mm for KF (see Figure 2 (c)).
**Average processing time for all three scenarios** was 115.1ms for EMICP, 27.03ms for WICP, 1.08ms for Horn, 22.8ms for RF and 10.43ms for KF (see Figure 2 (d), (e), (f)).

As illustrated in Table 1, one can obviously notice how significantly poorly EMICP and Horn perform. This drawback often occurs when the shapes present some symmetry. On the other hand, WICP is better endowed to cope with such drawbacks since it leverages knowledge about the quality of features, which helps it in discarding noisy elements. More importantly, KF and RF are both comparably superior in term of accuracy, but RF is more precise due to the control of uncertainty in parameters.

# Conclusion & Future Works

A novel approach for robust 3D point cloud registration was presented. This contribution is based on a recursive optimal state estimation. After establishing the link between WLS and its original counterpart (LS), 3D point cloud registration problem was fitted to KF scheme. However, since KF parameters for 3D registration (state and projection matrices) are built from noisy data, a non-negligible estimation instability was noticed. Consequently, we modelled the uncertainty and overcame it with an RF-based solution.

The accuracy of the proposed solution was tested on many synthetic as well as real 3D samples delivered by Kinect. Precision on the other hand, can be seen on the relatively small difference in accuracy among comparably noisy samples (red error bars in Figure 1 (c, d), Figure 2 (a, b, c) on the black line).

The proposed solution requires some feature points to be extracted from the source and the target point clouds before the alignment is carried out. The number of keypoints is relatively small compared to the size of point clouds. In addition, our solution can be extended to any dimension for data that can be point clouds, meshes as well as surfaces, given that some distinctive features are available.

As a future work, we intend to investigate alternative applications of recursive filtering algorithms in the field of computer vision. It would be also interesting to implement RF registration in the graphic processor to reach higher frame rates. In addition, in a multiview registration scenario (many sensors streaming images concurrently), data fusion algorithms open a new perspective for the users to reconstruct 3D scenes and to track moving objects cooperatively. This new horizon is convenient for the technologies of virtual and augmented reality.

---

[1] http://www.microsoft.com/en-us/kinectforwindows/. 2015

[2] http://www.naturalpoint.com/optitrack/. 2015

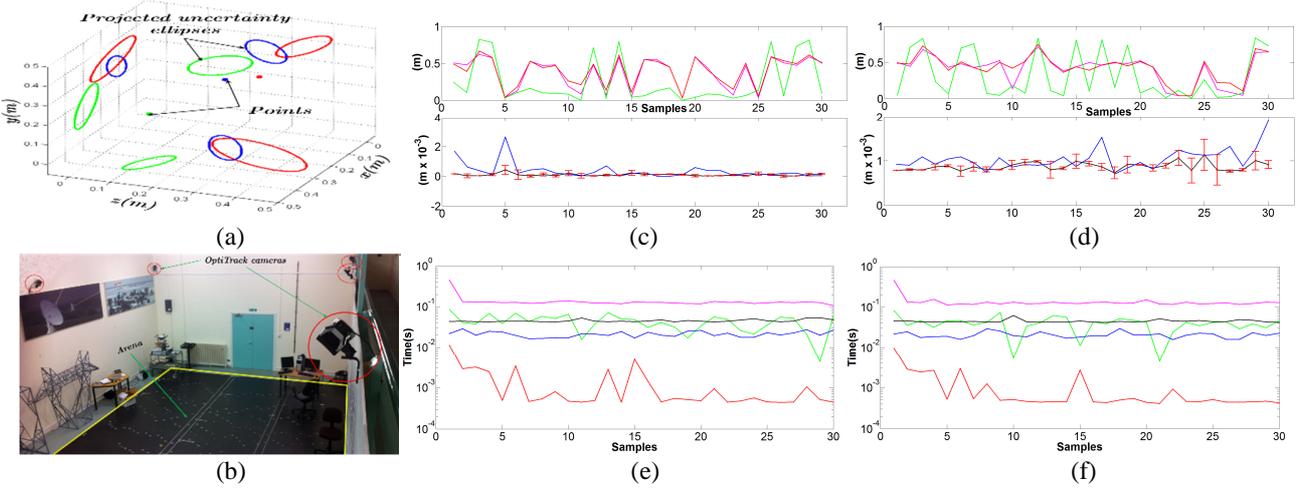

Figure 1 (a) 3D points uncertainty ellipses, (b) Ground truth (OptiTrack) and real data acquisition with Kinect in an indoor scene, (c), (d) 3D registration RMSE(m) of the New and the Old Kinect, respectively; (e), (f) Time elapsed during registration for the New and the Old Kinect, respectively. *EMICP (pink), WICP (green), Horn (red), RF (black), KF (blue)*

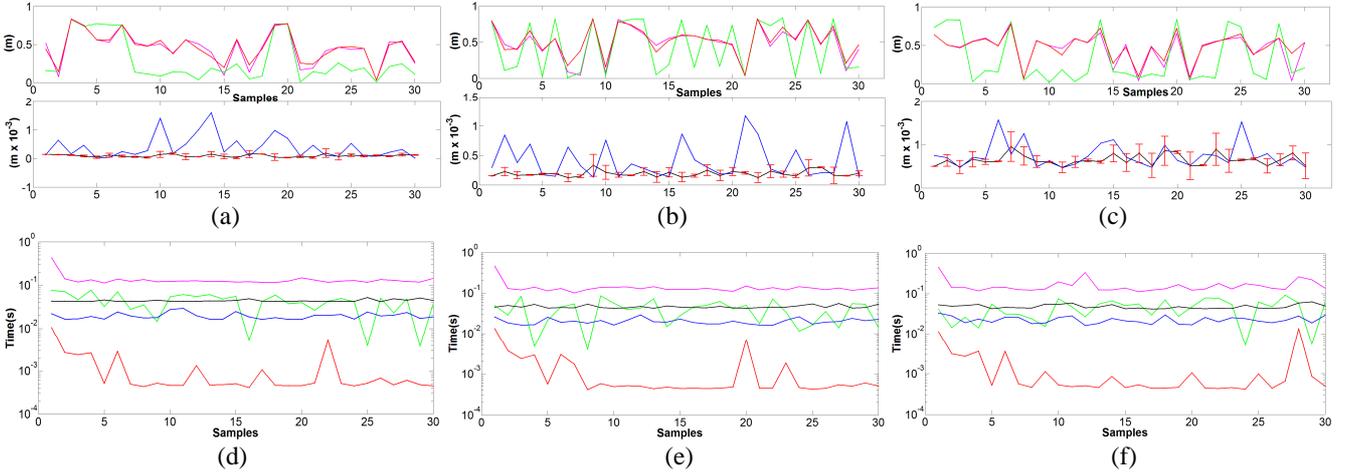

Figure 2 (a), (b), (c) 3D registration RMSE of the Small, Average and Large noise, respectively; (d), (e), (f) Time elapsed during registration for Small, Average and Large noise, respectively. *EMICP (pink), WICP (green), Horn (red), RF (black), KF (blue)*

| Noise | | EMICP | WICP | Horn | RF | KF |
|---|---|---|---|---|---|---|
| *New* | Kinect | 274 | 152 | 246 | 0.72 | 1.62 |
| *Old* | | 310 | 162 | 302 | 1.03 | 2.03 |
| *Small* | | 298 | 193 | 285 | 0.55 | 1.52 |
| *Average* | | 323 | 235 | 315 | 0.91 | 1.78 |
| *Large* | | 332 | 260 | 343 | 0.96 | 2.10 |

Table 1 RMSE (mm) for the whole sets of samples: 1000 for each simulation scenario and 120 for every version of Kinects